%% file: aaai2026.tex
\documentclass[letterpaper]{article} 

\usepackage{aaai2026}  
\usepackage{times}  
\usepackage{helvet}  
\usepackage{courier}  
\usepackage[hyphens]{url}  
\usepackage{graphicx} 
\usepackage{natbib}  
\usepackage{caption} 
\usepackage{algorithm}
\usepackage{algorithmic}

\usepackage{xcolor}
\definecolor{darkgreen}{rgb}{0,0.5,0}
\usepackage{soul}
\usepackage{pifont}
\usepackage{multirow}
\usepackage{booktabs}
\usepackage{amssymb}
\usepackage{subcaption}
\usepackage[most]{tcolorbox}
\usepackage{tabularx}
\usepackage{url}
\usepackage{xspace}
\newtcolorbox{promptbox}{
  colback=gray!10!white, colframe=black, sharp corners,
  boxrule=0.3mm, top=10pt, bottom=10pt, left=10pt, right=10pt, breakable
}

\definecolor{mred}{RGB}{236,151,165}
\definecolor{mgreen}{RGB}{200,229,179}
\definecolor{myellow}{RGB}{252,241,210}
\definecolor{mblue}{RGB}{65,102,245}

\usepackage{newfloat}
\usepackage{listings}
\DeclareCaptionStyle{ruled}{labelfont=normalfont,labelsep=colon,strut=off} 
\floatstyle{ruled}
\newfloat{listing}{tb}{lst}{}
\floatname{listing}{Listing}
\newcommand{\ourdataset}{\texttt{FormulaQA}\xspace}
\newcommand{\ourdatasetda}{\texttt{FormulaQA(Direct)}\xspace}
\newcommand{\ourdatasetcot}{\texttt{FormulaQA(CoT)}\xspace}
\newcommand{\ourmethod}{\texttt{TabAF}\xspace}
\newcommand{\ourmethodda}{\texttt{TabAF-Fast}\xspace}
\newcommand{\ourmethodcot}{\texttt{TabAF-CoT}\xspace}

\title{General Table Question Answering via Answer-Formula Joint Generation}
\author{
    Zhongyuan Wang\textsuperscript{\rm 1},
    Richong Zhang\textsuperscript{1,2}\thanks{Corresponding author},
    Zhijie Nie\textsuperscript{1,3},
    Hangyu Mao\textsuperscript{4},
}
\affiliations{
    \textsuperscript{\rm 1}CCSE, School of Computer Science and Engineering, Beihang University, Beijing, China\\
     \textsuperscript{\rm 2}Zhongguancun Laboratory, Beijing, China\\
     \textsuperscript{\rm 3}Shen Yuan Honors College, Beihang University, Beijing, China\\
     \textsuperscript{\rm 4}Kuaishou Technology

}

\usepackage{bibentry}

\begin{document}
\nocopyright
\maketitle

\input{0-Abstract}
\input{1-Introduction}
\input{2-RelatedWorks}
\input{3-Dataset}

\input{4-Framework}

\input{5-Experiments}

\input{6-Conclusion}

\bibliography{aaai2026}

\appendix
\input{8-Appendix}

\end{document}

%% file: 0-Abstract.tex
\begin{abstract}
Advanced table question answering (TableQA) methods prompt large language models (LLMs) to generate answer text, SQL query, Python code, or custom operation, which impressively improve the complex reasoning problems in the TableQA task. However, these methods lack the versatility to cope with specific question types or table structures. In contrast, the Spreadsheet Formula, the widely used and well-defined operation language for tabular data, has not been thoroughly explored to solve TableQA. In this paper, we first attempt to use the Formula as the executable representation for solving complex reasoning on tables with different structures. Specifically, we construct \ourdataset, a large Formula-annotated TableQA dataset from existing datasets. In addition, we propose \ourmethod, a general table answering framework to solve multiple types of tasks over multiple types of tables simultaneously, which decodes answers and Formulas with a single LLM backbone. Extensive experiments demonstrate the versatility and generalization of \ourmethod. Under the same model size, \ourmethod achieves new state-of-the-art performance on the WikiTableQuestion, HiTab, and TabFact. \footnote{Our code and dataset is available at \url{https://github.com/inforin/TabAF}}
\end{abstract}

%% file: 1-Introduction.tex
\section{Introduction}
A table is an arrangement of information or data in rows and columns, which is ubiquitous and serves as an important data carrier in many domains. The Table Question Answering (TableQA) task aims to answer natural language questions based on tables, requiring models to comprehend and analyze tabular content. Unlike the database tables in the Text-to-SQL task, tabular data in TableQA exhibits more flexible structures and mixed data types, making it more challenging to handle with SQL queries. The questions in TableQA can be categorized into two types: lookup questions, which select part of the table contents as answers, and complex reasoning questions, which require calculating or aggregating multiple values from tables.

\begin{figure}[th]
    \centering
    \includegraphics[width=\linewidth]{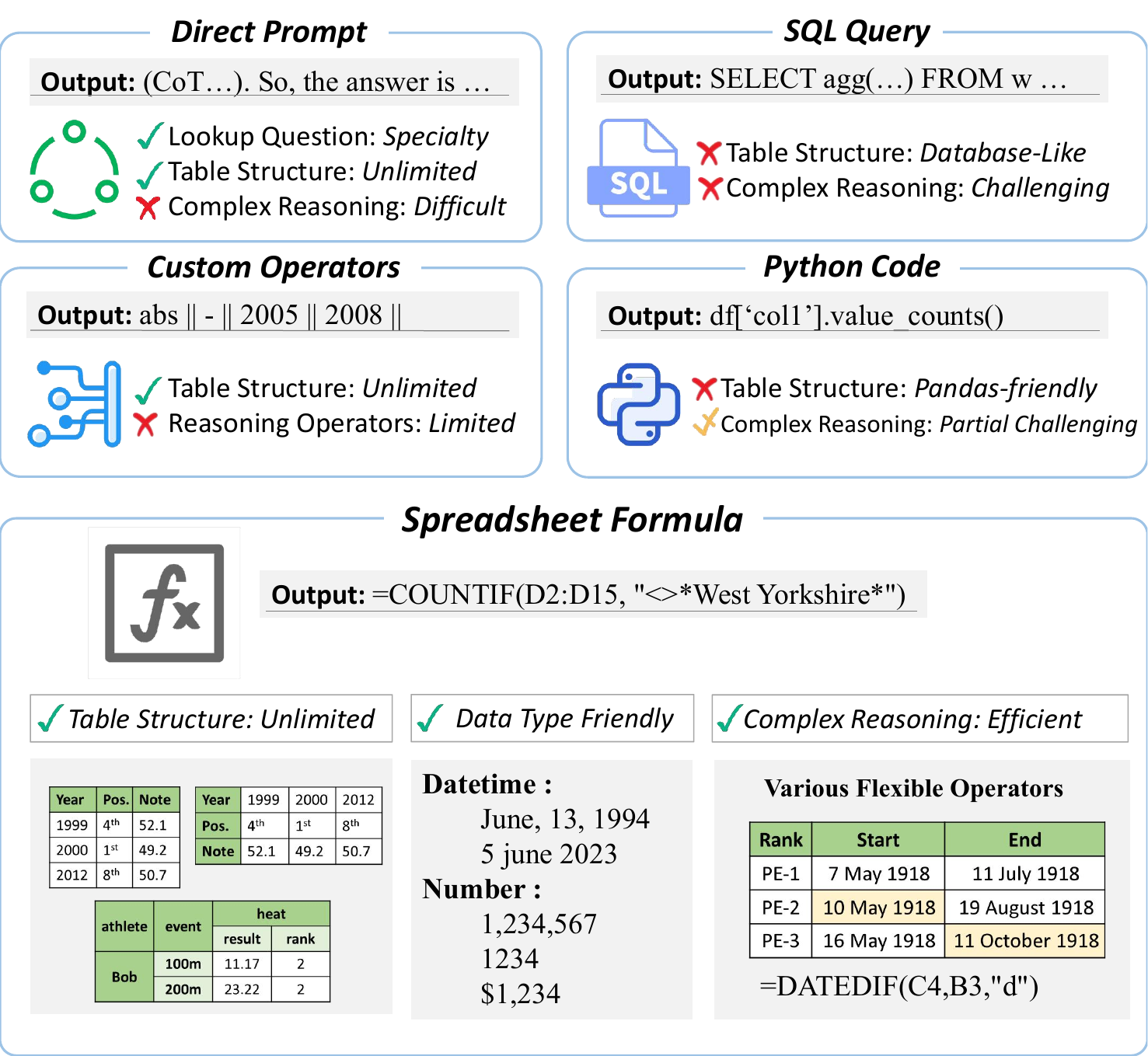}
    \caption{Different generation modes in TableQA.}
    \label{fig:introduction}
\end{figure}

Current mainstream TableQA methods leverage the capability of large language models (LLMs), either prompting LLMs to generate outputs without training \cite{cheng2023binding, wang2024chainoftable, yang2025triples} or fine-tuning LLMs to enhance table reasoning capabilities \cite{zhang2024tablellama, zhang2024tablellm}. In general, four generation modes are used in existing LLM-based TableQA methods: Direct Prompting (DP) mode aims at prompting LLMs to generate textual answers directly or to conclude the final answers after reasoning through Chain-of-Thought (CoT) \cite{liu-etal-2024-rethinking,jiang2024seek}, which are highly flexible and are not restricted by table structures. The DP-based methods excel at looking up existing answers from tables, particularly when extracting substring content from complex cells. The remaining three are the generation modes via executable representations, including generating SQL queries \cite{cheng2023binding, level}, Python codes \cite{su2024tablegpt2, yang2025triples}, and custom operations  \cite{mouravieff2024learning}. These modes generate executable representations and invoke external engines to derive answers, which are more adept at complex reasoning. 

However, as illustrated in Figure \ref{fig:introduction}, existing generation modes all have non-negligible limitations: (1) The DP mode relies on the intrinsic reasoning capability of LLMs, struggling to solve complex reasoning over tables;(2) The SQL-based methods assume tables are database-like, making it difficult to apply to hierarchical tables \cite{squall}; (3) The Python mode often struggles with complex table structures. For instance, the commonly used third-party toolkit \texttt{Pandas}\footnote{\url{https://pandas.pydata.org/}} mishandles horizontal tables and requires extra steps for mixed data processing and multi-index lookup \cite{liu-etal-2024-rethinking}, increasing code complexity; (4) The methods generating custom operations define a limited set of reasoning operators, exhibiting poor scalability. 

Therefore, a crucial question arises: \textit{\bf Whether there is a generalized executable representation that can deal with various questions on the diverse tables?} In this paper, we explore the potential of one promising candidate, Spreadsheet Formulas (abbreviated as Formulas). As the native operation language of tabular data, Formulas provide rich operators specifically designed for tabular data. Prior work \cite{cheng-etal-2022-fortap} shows its potential on simple arithmetic reasoning over tables, overlooking its general question-answering capabilities. Meanwhile, generating Formulas via LLMs for TableQA has not been explored. 

We believe that generating Formulas offers the following advantages: (1) \textbf{Flexible:} Similar to DP-based methods, Formulas can adapt to tables with different structures, such as horizontal tables, hierarchical tables, or tables with mixed data types. (2) \textbf{Data-Friendly:} The Formula execution engine is compatible with different formatted values, such as formatted numbers or dates, with no need for data type transferring; (3) \textbf{Efficient:} Formulas contain rich table operators, enabling complex reasoning with concise expressions, which are difficult for Python or SQL to achieve. 

To unleash the potential of LLMs in generating Formulas, {\bf we constructed \ourdataset, a large Formula-annotated dataset for TableQA.} Specifically, we first develop \textbf{\ourdatasetda}, a large-scale dataset tailored for direct Formula generation, aimed at enhancing the ability of LLMs to generate task-specific Formulas for TableQA. \ourdatasetda is through a combination of existing Formula data collection, SQL-to-Formula conversion, and annotations by a strong LLM. It contains 24,738 samples, each with complete (question, table, answer) triplets and corresponding Formulas. Furthermore, we construct \textbf{\ourdatasetcot}, a small-scale reasoning-focused subset with high-quality CoT annotations by Claude-Sonnet-4, serving as cold start data for RL fine-tuning of \ourmethodcot, which encourages LLMs to think before generating answers or Formulas.

Both DP and Formula modes are well-suited for different tables, with distinct problem-solving focuses. The DP mode is tailored for lookup questions, while the Formula mode excels at handling complex reasoning questions. To combine their advantages, {\bf we propose \ourmethod, an answer-Formula joint generation and aggregation framework}, which supports both DP and Formula generation modes by producing outputs from both modes and selecting the final answer to a question via majority voting. \ourmethod comprises two variants: \textbf{\ourmethodda}, which directly generates answers or Formulas based on the question and table without an explicit reasoning process, balancing performance and efficiency; and \textbf{\ourmethodcot}, which generates answers or Formulas after a CoT reasoning process, aiming for higher performance. To fully exploit the potential of LLMs, we first perform supervised fine-tuning (SFT) to both variants, followed by a reinforcement learning (RL) training stage. 

Based on \ourdataset, we trained \ourmethod on Qwen2.5-Coder-7B \cite{hui2024qwen2} and Llama3.1-8B \cite{dubey2024llama} using \ourdataset. {\bf Extensive experiment analysis shows the superiority of our approach}: for in-domain setting, \ourmethodda significantly surpasses fine-tuning-based baselines on WTQ, HiTab, and TabFact while maintaining high token efficiency; \ourmethodcot demonstrates stronger reasoning ability through CoT generation; for out-of-domain setting, both variants demonstrate generalization capabilities on AIT-QA for aviation-field tableQA and FinQA for finance-field HybridQA, achieving substantially better performance than existing TableQA LLMs; Ablation studies demonstrate the necessity of both modes in \ourmethod and the effectiveness of each training stage in our methods.

%% file: 2-RelatedWorks.tex
\section{Related Work}
\paragraph{Prompting-based TableQA Methods}
Most LLM-based methods leverage the in-context learning ability of LLMs to predict answers or executable representations such as SQL or Python code. Binder \cite{cheng2023binding} prompts LLMs to generate SQL queries. Chain-of-Table \cite{wang2024chainoftable} and API-Assisted \cite{cao2023api} design Python functions to filter flat and hierarchical tables, respectively. Recently, \cite{liu-etal-2024-rethinking} and \cite{yang2025triples} perform joint reasoning over multiple outputs from different methods, achieving the best performance on flat tables. However, most existing methods are limited to either flat or hierarchical tables and lack generalizability across diverse table structures. Furthermore, many rely on closed-source models such as ChatGPT for generation.

\paragraph{Fine-tuning-based TableQA Methods}
Traditional methods  \cite{yin2020tabert, herzig2020tapas} utilize data with different input or output formats during pretraining to enhance the reasoning ability of small models. FORTAP \cite{cheng-etal-2022-fortap} first introduces Formulas into the TableQA task, training models with Formulas annotated in Hitab \cite{cheng2022hitab}, where the Formulas are limited to arithmetic operations. Several recent studies, including TableLLama \cite{zhang2024tablellama}, TableLLM \cite{zhang2024tablellm}, TableBench \cite{wu2024tablebench}, and TableGPT2 \cite{su2024tablegpt2}, attempt to improve the table reasoning ability of LLMs by fine-tuning on more tabular data. Their fine-tuned LLMs struggle to perform well on a single task. 

\paragraph{SpreadSheet Formula Prediction}
The Formula prediction task aims to complete or derive spreadsheet content without relying on natural language questions, which \textbf{differs from TableQA}. SpreadSheetCoder \cite{chen2021spreadsheetcoder} completes Formulas for cells in given tables. Auto-Formula  \cite{chen2024auto} recommends Formulas for cells based on similar tables. In contrast, NL2Formula \cite{zhao2024nl2Formula} provides a benchmark for generating verbose and inefficient formulas for database-like tables, which differ significantly from the tabular data addressed in TableQA.

%% file: 3-Dataset.tex
\begin{table*}[th]
    \centering
    \small
    \resizebox{\linewidth}{!}{
     \begin{tabular}{l|l|l}
        \toprule
        \bf Type & \bf Formula & \bf SQL Template Example \\
        \midrule
        \multirow{2}{*}{Lookup} & =F28 & SELECT c4 FROM w WHERE c3 = `jaime quintana' \\
        & =B39 & SELECT c2 FROM w ORDER BY id DESC LIMIT 1 \\
        \midrule
        & =MIN(G2:G35) & SELECT MIN(c3) FROM w WHERE agg = 0 \\
        & =COUNTA(A2:A36) & SELECT COUNT(*) FROM w WHERE agg = 0 \\
        Complex & =COUNTA(UNIQUE(A2:A39)) & SELECT COUNT(DISTINCT c1) FROM w WHERE agg = 0 \\
        Reasoning & =SUMIFS(N2:N22, G2:G22,``=60'' & SELECT SUM(c4\_number) FROM w WHERE c2\_year = 60 \\
        & =MINIFS(D2:D33, G2:G33,``=a'', H2:H33,``<>b'') & SELECT MIN(c2) FROM w WHERE c3 = `a' AND c4 != `b' \\
        & =INDEX(A2:A13, MATCH(MIN(D2:D13),D2:D13,0)) & SELECT c1 FROM w ORDER BY c2\_parsed ASC LIMIT 1 \\
        \bottomrule
    \end{tabular}
    }
    \caption{Examples of some SQL templates we utilize for Formula conversion. }
    \label{tab:transfer_sql_examples}
\end{table*}

\section{FormulaQA}
Spreadsheet Formulas are a native and widely used tool for handling tabular data. However, open-source LLMs often struggle with Formula reasoning due to limited exposure during training, especially for Formulas tailored to TableQA tasks. To enhance the Formula generation capability for LLMs and leverage the advantages of Formulas in accomplishing complex reasoning over different structured tables, \textbf{we first construct \ourdatasetda, a large-scale spreadsheet dataset}, enabling the LLMs to reestablish proficiency in Formula generation for TableQA. \ourdatasetda avoids explicit reasoning steps, enabling efficient use of existing data and reducing CoT annotation overhead. To further elicit the reasoning capability, \textbf{we introduce \ourdatasetcot, a small-scale CoT dataset used after extensive training on \ourdatasetda}, to guide the LLMs toward a reasoning-then-generation paradigm, preparing them for subsequent RL training. \textbf{We do not construct a separate test set for either variants }, as we only reformulate the train and dev sets from existing TableQA datasets. Their original test sets can be directly used for evaluation.

\subsection{FormulaQA(Direct)}
To ensure our Formula dataset covers common TableQA scenarios, we first collect existing Formula data in QA domains. Then, to address the insufficient Formula data from existing datasets and expand the basic complex reasoning capability, we utilize the SQL-solvable subset within TableQA by converting SQL queries into Formulas via templates. Finally, we annotate Formulas using the powerful LLM for SQL-unsolvable or SQL-intractable questions.

\paragraph{Existing Formula Collection} We collect Formulas from HiTab \cite{cheng2022hitab}, as it is the only TableQA dataset with Formula annotations. The Formulas in HiTab are relatively short, mainly involving assignment Formulas (e.g., ``$\tt =B2$'') and basic arithmetic Formulas, which ensures the fundamental capability for reasoning. 

\begin{table}[t]
    \centering
    \small
    \resizebox{\linewidth}{!}{
    \begin{tabular}{l|cccc}
        \toprule
        \multirow{2}{*}{\bf Dataset} & \multicolumn{3}{c}{\bf FormulaQA(Direct)} & \bf FormulaQA \\
        & Train & Dev & Avg Token & \bf (CoT) \\
        \midrule
        HiTab & 7,409 & 1,671 & 3.45 & 1,200 \\
        Squall \& WTQ & 9,829 & 2,439 & 19.35 & 1,200  \\
        TabFact & 7,500 & 1,800 & 36.49 & 1,200  \\
        \bottomrule
    \end{tabular}
    }
    \caption{Statistics of Formulas dataset, the average token count is calculated using the Qwen tokenizer.
    }
    \label{tab:corpus_stats1}
\end{table}

\paragraph{Templated-Based SQL to Formula} We collect the SQL data annotated for TableQA to convert into Formulas. WTQ \cite{wtq} is a mainstream challenging TableQA benchmark. Squall \cite{squall} standardizes and extends WTQ tables, annotating SQL queries and associated templates for SQL-solvable questions. We count their SQL templates and manually write conversion programs for common templates. We provide example SQL queries and the corresponding Formulas for partial templates in Table \ref{tab:transfer_sql_examples}, and explain the conversion details in Appendix A. 

\paragraph{Annotation by the Powerful LLM} 
Since the remaining long-tail templates in Squall are difficult to convert, and many SQL-unsolvable questions can still be expressed using Formulas, we annotate the remaining WTQ questions using GPT-4o. For each question, we prompt the LLM to generate 10 Formulas, execute them, and select the most concise one with correct results. If no Formula executes correctly, back the incorrect outputs for new Formula generation, up to 3 attempts per question. We also randomly select samples from TabFact, a fact-checking dataset commonly used in TableQA, to enhance the variety of Formula. The prompt used for annotation can be found in Appendix D.1.

\subsection{FormulaQA(CoT)}
To ensure a high-quality cold start for subsequent RL training, we employ Claude-Sonnet-4, the SOTA reasoning model, to annotate CoT-style reasoning paths. Specifically, we randomly sample examples from the train sets of WTQ, HiTab, and TabFact, prompting the powerful LLM to think before generating answers or Formulas. Each question allows up to three annotation attempts; only outputs with correct answers or Formula-executed results are retained, while failed samples are discarded. The prompt used for CoT annotation can be found in Appendix D.2.

\begin{figure*}[ht]
    \centering
    \includegraphics[width=0.9\linewidth]{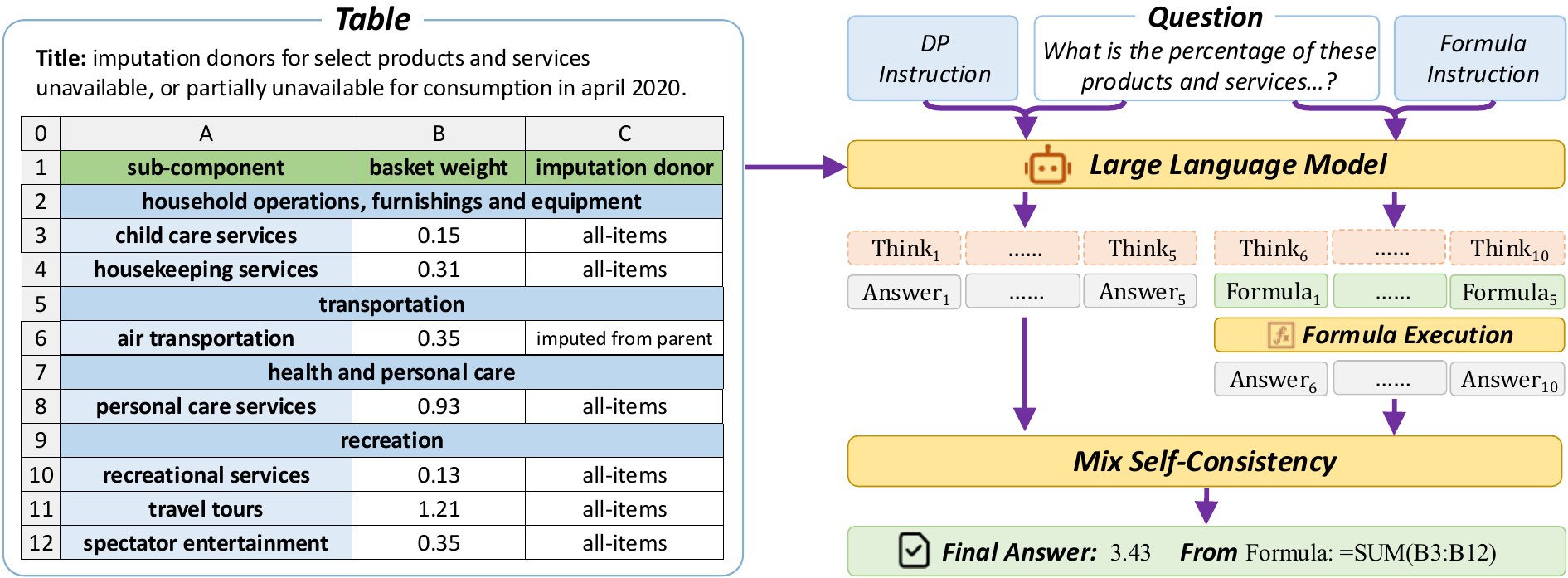}
    \caption{The overall structure of \ourmethod. The models are trained using \ourdataset\ and existing tabular data. We prompt the models with different instructions to generate Formulas and answers, where thinking is optional. The mix self-consistency strategy is employed to obtain the final answer from the Formula-executed answers and the DP-based generated ones.}
    \label{fig:framework}
\end{figure*}

\subsection{Dataset Statistic}

The statistics of \ourdataset are summarized in Table \ref{tab:corpus_stats1}. For \ourdatasetda, we follow the original dataset splits to construct the train and dev sets. We also report the token statistics of Formulas from different datasets. Formulas in Hitab consist of only a few tokens, mainly including lookup and simple arithmetic Formulas. Formulas for WTQ and Squall are longer, containing more complex reasoning operators such as filtering, sorting, and aggregations. Formulas for TabFact are even longer, as the fact-checking task requires Formulas to perform verifications across multiple conditions. For \ourdatasetcot, we randomly sample and annotate 600 CoT examples for Formula generation and 600 for answer generation from the train set of each dataset.

%% file: 4-Framework.tex
\section{TabAF}
We introduce \ourmethod, an answer-Formula joint generation framework for TableQA that combines the strengths of Formula mode in complex reasoning and DP mode in lookup questions. As shown in Figure \ref{fig:framework}, given a new question, \ourmethod is guided by mode-specific instructions to generate 5 Formulas and 5 answers. Following previous works \cite{liu-etal-2024-rethinking}, we employ the mix self-consistency strategy to determine the final answer from 10 candidates.  

To meet practical demands in real-world scenarios, we develop two variants of \ourmethod: \textbf{\ourmethodda prioritizes inference efficiency by directly generating answers or Formulas, whereas \ourmethodcot emphasizes performance by incorporating a reasoning process before generating answers or Formulas.}
The two-generation modes of both variants of \ourmethod can be expressed as
\begin{equation}
\begin{aligned}
   & f_{\rm LLM}(I_{\rm DP}, T, Q)\to [R, ]\,Y_{\rm DP} \\
   & f_{\rm LLM}(I_{\rm Formula}, T_{\rm ex}, Q)\to [R, ]\,Y_{\rm Formula}
\end{aligned}
\end{equation}
where $Y_{\rm DP}$ and $Y_{\rm Formula}$ are the output answer and Formula, separately. $R$, included exclusively in \ourmethodcot, denotes the reasoning process before generating the final output. $I_{\rm DP}$ and $I_{\rm Formula}$ refers to the instructions of different modes, $Q$ refers to the question. $T$ refers to the origin table in markdown format, while $T_{\rm ex}$ refers to the augmented view of the table $T$, which is added with column labels (A, B, C...) and row numbers (1, 2, 3, ...) of the spreadsheets to assist LLMs in generating Formulas. The prompts for these two variants can be found in Appendix D.3 and D.4.

For Formula execution, due to the limited usability of the Excel engine on Linux, we use the open-source package \texttt{formulas} \footnote{\url{https://github.com/vinci1it2000/formulas}} for efficient batched Formula execution. As this package is relatively lightweight, it does not fully match the Excel engine in terms of the number and functionality of supported Formulas. To address this, we extended its Formula coverage based on the needs observed in Formula annotations to better simulate the behavior of the Excel engine.

\subsection{Training Pipeline}
To simultaneously enhance the generation capabilities of LLMs under both generation modes, fine-tune the LLMs in a multi-task learning approach, where LLMs are guided by mode-specific instructions to produce the expected outputs. In terms of the training pipeline, for \ourmethodda, which directly generates answers or Formulas, we first perform supervised fine-tuning (SFT), followed by reinforcement learning (RL). For \ourmethodcot, we introduce an additional cold-start stage before RL training, enabling LLMs to internalize the CoT style reasoning. 

\paragraph{Supervised Finetuning Stage}
To familiarize the LLMs with the downstream TableQA task and the use of Formulas within it, we train LLMs on \ourdatasetda and the (question, table, answer) triplets from the original TableQA datasets, which include the training splits of WTQ and HiTab, and a sampled subset of TabFact for balance.

\paragraph{Cold Start Stage for \ourmethodcot}
The model trained in the previous SFT stage is only capable of directly generating Formulas or answers. Therefore, before RL training, we warm up the model using a small amount of CoT-style data to activate its ability to produce CoT-style outputs. In this stage, we use \ourdatasetcot for training, which contains CoT examples for both the DP mode and Formula mode, sampled from the same datasets as in SFT.

\paragraph{RL Training Stage} 
We further fine-tune the LLMs using reinforcement learning with Group Relative Policy Optimization (GRPO) \cite{shao2024deepseekmath}. To avoid introducing new datasets or scaling data volume, we continue to use the full train sets of WTQ and HiTab, along with the subset of TabFact. LLMs are guided by specific instructions to generate either answers or Formulas. Rewards are assigned based on the correctness and format consistency of the generated outputs. In terms of reward function design:
\begin{itemize}
\item For \ourmethodda, since it directly generates direct answers or Formulas, only correctness-based rewards are needed. The reward function is defined as:
  \[
R = 
\begin{cases}
1, & \text{if } \text{Correct}(Y) = 1 \\
0, & \text{otherwise}
\end{cases}
\]

where $\text{Correct}(Y)$ indicates whether the output $Y$ (either a predicted answer or an executed Formula result) matches the ground truth.

\item For \ourmethodcot, which generates intermediate reasoning steps before producing the final output, we further require the response to follow a specific format—namely, a \texttt{<think>...</think>} reasoning segment followed by a mode-specific \texttt{<mode\_tag>...</mode\_tag>} output. Therefore, the reward function jointly considers both format correctness and answer correctness:

\[
R = 
\begin{cases}
1.5, & \text{if } \text{Format}(R, Y) = 1 \land \text{Correct}(Y) = 1 \\
0.5, & \text{if } \text{Format}(R, Y) = 1 \land \text{Correct}(Y) = 0 \\
0,   & \text{if } \text{Format}(R, Y) = 0
\end{cases}
\]

where $\text{Format}(R, Y)$ indicates whether the reasoning process $R$ and the final output $Y$ adhere to the format.
\end{itemize}

%% file: 5-Experiments.tex
\section{Experiments}
\subsection{Experiment Setup}
\paragraph{In-Domain Datasets} We evaluate our method on 
WikiTableQuestion (WTQ) \cite{wtq}, HiTab \cite{cheng2022hitab}, and TabFact \cite{Chen2020TabFact:}, with the corresponding training sets included in the training data. WTQ is the most commonly used TableQA benchmark, with 4,344 test samples. It contains complex questions that require reasoning across multiple entities in the given table to obtain the answers. HiTab is a hierarchical web table benchmark derived from statistical reports and Wikipedia pages, where the tables usually contain rich table structures and numerical values. The test set has 1,584 samples. TabFact is a table-based fact verification dataset that determines whether a statement matches a given table. We evaluate on its test-small set (2,024 samples). We also divide the complex subsets for WTQ and HiTab (with 2,421 and 217 examples, respectively) to evaluate the complex reasoning capabilities, and the division criteria are illustrated in Appendix B.

\paragraph{Out-of-Domain Dataset}
We choose AIT-QA \cite{katsis2022ait} and FinQA \cite{chen2021finqa} as out-of-domain datasets to evaluate the generalization ability of \ourmethod. AIT-QA is a hierarchical table benchmark, containing 515 samples with 113 tables from airline domains. FinQA is a hybrid QA dataset in the financial domain, which requires numerical reasoning over tables and the surrounding text to obtain answers. Its public test set contains 1,147 samples.

\paragraph{Evaluation Metrics}
Following previous work \cite{cheng2023binding, dater}, we employ denotation accuracy as the evaluation metric for WTQ and utilize the official evaluation script. For TabFact, we use binary classification accuracy, and for other datasets, we use accuracy. We also report the average tokens generated (\#Token) on WTQ to illustrate the efficiency of \ourmethodda. 

\paragraph{LLMs}
We apply \ourmethod to two widely used LLMs: Qwen2.5-Coder-7B \cite{hui2024qwen2}, and Llama3.1-8B \cite{dubey2024llama}. During the training stage, we utilize LlamaFactory \cite{zheng-etal-2024-llamafactory} for SFT and verl \cite{sheng2025hybridflow} for RL training. During inference, we utilize vLLM \cite{kwon2023efficient} for efficient LLM generation. All experiments are conducted on 8 NVIDIA GPUs with performance no less than that of the A100 40GB or 80GB. All hyperparameters can be found in Appendix C.

\begin{table*}[t]
    \centering
    \small
    \resizebox{\linewidth}{!}{

    \begin{tabular}{l|l|c|ccccc}
        \toprule
        \multirow{2}{*}{\textbf{Method}} & \multirow{2}{*}{\textbf{Backbone}} & \multirow{2}{*}{\textbf{\#Token}} & \multicolumn{2}{c}{\textbf{WTQ}} & \multicolumn{2}{c}{\textbf{HiTab}} & \multirow{2}{*}{\textbf{TabFact}} \\ 
         & & & Comp. & ALL & Comp. & ALL & \\ 
        \midrule
        \multicolumn{8}{c}{\textit{Prompting-Based Methods}} \\
        \midrule
        API-Assisted  \cite{cao2023api} & CodeX & - & - & 42.40 & - & 69.30 & -  \\
        ReAcTable  \cite{reactable} & CodeX & - & - & 68.00 & - & - & 86.10 \\
        Chain-of-Table  \cite{wang2024chainoftable} & PaLM 2 & - & - & 67.31 & - & - & 86.61  \\
        Norm-DP\&Agent \cite{liu-etal-2024-rethinking} & GPT-3.5 & - & - & 73.65 & - & - & 88.50  \\
        TIDE DP\&Agent \cite{yang2025triples} & GPT-3.5 & - & - & 75.00 & - & - & 89.82   \\
        E5 \cite{zhang-etal-2024-e5} & GPT-4 & - & - & - & - & \textbf{85.08} & -  \\
        \midrule
        \multicolumn{8}{c}{\textit{Fine-tuning-Based Methods}} \\
        \midrule
        TAPEX-Large  \cite{liu2022tapex} & BART-Large & - & - & 59.10 & - & - & 84.20  \\
        TableLlama \cite{zhang2024tablellama} & Llama2-7B & 7.47 &  &  & 22.12 & 60.48 & 82.55  \\
        TableLLM(TCoT) \cite{wu2024tablebench} & Qwen2-7B & 364.40 & 45.77 & 53.59 & 12.9 & 43.88 & 69.81  \\
        TableLLM(TCoT) \cite{wu2024tablebench} & Llama3.1-8B & 320.84 & 44.28 & 50.18 & 9.22 & 44.7 & 30.29  \\
        TableGPT2 \cite{su2024tablegpt2} & Qwen2.5-7B & 116.36 & - & 61.42 & - & 70.27 & 77.80  \\
        \midrule
        \multirow{2}{*}{\ourmethodda} 
        & Qwen2.5-Coder-7B  & 109.13* & 73.36 & 76.10 & \bf 60.83 & 78.03 & 85.03 \\
        & Llama3.1-8B       & 78.08* & 69.02 & 74.26 & \underline{60.37} & 79.99 & 83.00 \\
        \midrule
        \multirow{2}{*}{\ourmethodcot} 
        & Qwen2.5-Coder-7B  & 4257.80* & \underline{80.92} & \underline{82.94} & 54.38 & 80.49 & \underline{93.23} \\
        & Llama3.1-8B       & 4646.09* & \bf 82.07 & \bf 83.79 & 58.53 & \underline{81.69} & \bf 94.91 \\
        \bottomrule
    \end{tabular}
    }
    \caption{In-Domain performance of \ourmethod\ on WTQ, HiTab and TabFact, ``Comp.'' refers to the complex reasoning subset. ``\#Token'' refers to the average tokens generated on WTQ. ``*'' denotes the total number of tokens across 10 outputs. }
    \label{tab:main_results}
\end{table*}

\paragraph{Baselines}
We compare our framework with a wide range of baselines, including both prompting-based methods and fine-tuning-based methods. Prompting-based methods are the mainstream Table approaches, most of which rely on large, closed-source models. We compare our approach with these methods: API-Assistied \cite{cao2023api}, ReAcTable \cite{reactable}, Chain-of-Table \cite{wang2024chainoftable}, Norm \cite{liu-etal-2024-rethinking}, TIDE \cite{yang2025triples}, E5 \cite{zhang-etal-2024-e5}. For the fine-tuning-based methods, we choose LLMs that are fine-tuned or developed for the TableQA task, which includes TableLlama \cite{zhang2024tablellama}, TableLLM \cite{wu2024tablebench}, TableGPT2 \cite{su2024tablegpt2}. For a fair comparison, we only report their in-domain results on WTQ, HiTab, and TabFact.

\subsection{Main results}

\subsubsection{In-Domain Performance}\label{sec:in-domain}
\ul{\bf \texttt{TabAF} consistently outperforms existing methods across both variants.} As shown in Table \ref{tab:main_results}, Both variants of \ourmethod\ based on Qwen2.5-Coder-7B and Llama3.1-8B achieves significant performance on three datasets compared to other fine-tuning-based methods. The performance-oriented \ourmethodcot surpasses all other methods with a margin of at least 6.14\%, 3.41\% on WTQ, TabFact, respectively. On HiTab, it outperforms all methods except E5, which is specifically designed for hierarchical tables and uses the powerful closed-source LLM. On the complex reasoning subset of HiTab, \ourmethodda outperforms \ourmethodcot, possibly because the latter overthinks or overcomplicates the numerical reasoning questions in HiTab.
\\
\ul{\bf \texttt{TabAF-Fast} exhibits inference token efficiency.}
Since \ourmethodda generates direct Formulas and answers without thinking, \ourmethodda\ consumes the least number of tokens compared to methods that utilize Chain-of-Thought, or multi-stage iterative reasoning, indicating that \ourmethod can achieve fast and accurate responses in practical scenarios. Efficiency experiments show that \ourmethodda takes 57 seconds to perform inference on all test sets using 8 H800 GPUs, achieving 8.4× faster inference speed compared to \ourmethodcot, which takes 481 seconds. 
\\
\ul{\bf \texttt{TabAF} shows versatility across different table scenarios.}
Most baseline methods are designed for specific table scenarios and achieve optimal results on a single dataset. In contrast, our method consistently outperforms different specialized methods across multiple datasets, further highlighting the universality of the DP and Formula approach in the table reasoning task and the effectiveness of \ourmethod. 

\subsubsection{Out-of-Domain Performance}\label{sec:out-of-domain}
\begin{table}[t]
    \centering
    \small
    \resizebox{\linewidth}{!}{
    
    \begin{tabular}{l|l|cc}
        \toprule
        \textbf{Method} & \textbf{Backbone} & \textbf{AIT-QA} &  \textbf{FinQA} \\ 
        \midrule
        TableLlama & Llama2-7B & 26.99 & 2.27 \\
        TableLLM(TCoT) & Qwen2-7B & 64.85* & 8.63 \\
        TableLLM(TCoT) & Llama3.1-8B & 60.39* & 6.63 \\
        TableGPT2 & Qwen2.5-7B & 12.43 & 40.28 \\
        \midrule
        \multirow{2}{*}{\ourmethodda} 
        & Qwen2.5-Coder-7B  & 79.03 & 50.65 \\
        & Llama3.1-8B       & 79.81 & 47.95 \\
        \midrule
        \multirow{2}{*}{\ourmethodcot} 
        & Qwen2.5-Coder-7B  & \underline{87.18} & \bf 60.68 \\
        & Llama3.1-8B       & \bf 88.74 & \underline{57.80} \\
        \bottomrule
    \end{tabular}
    }
    \caption{Out-of-domain performance of \ourmethod\ on AIT-QA and FinQA. ``*'' indicates partially in-domain performance.}
    \label{tab:ood_results}
\end{table}
As shown in Table \ref{tab:ood_results}, \ourmethod maintains performance on both out-of-domain datasets. For AIT-QA, the hierarchical table dataset, our Formula execution is not limited by the table structure, whereas TableGPT-2 struggles to execute its Python code successfully on such raw hierarchical tables. For FinQA, the HybridQA dataset that requires reasoning questions on both tables and texts associated with tables, \ourmethod, despite not being specifically fine-tuned on the HybridQA task, can still leverage its ability to perform complex reasoning like numerical calculation on the FinQA dataset by Formula generation, outperforming other fine-tuning-based methods. The DP-based methods, TableLlama and TableLLM (TCoT), struggle with such complex reasoning questions. TableGPT2 benefits from the ability to generate Python code to solve some complex reasoning questions, but still exhibits a high code error rate in HybridQA scenarios. 

\subsection{Ablation Study}\label{sec:ablation}

\begin{table}[t]
    \centering
    \small
    \resizebox{0.9\linewidth}{!}{
    \begin{tabular}{l|ccccc}
        \toprule
        \multirow{2}{*}{\textbf{Method}} & \multicolumn{2}{c}{\textbf{WTQ}} &  \multicolumn{2}{c}{\textbf{HiTab}} & \multirow{2}{*}{\textbf{TabFact}} \\ 
         & Comp. & ALL & Comp. & ALL & \\ 
        \midrule
        \multicolumn{6}{c}{\textit{\ourmethodda}} \\
        \midrule
        Upper Bound & \textcolor{gray}{82.69} & \textcolor{gray}{85.20} & \textcolor{gray}{67.28} & \textcolor{gray}{84.22} & \textcolor{gray}{96.00} \\
        DP\&Formula     & 73.36 & 76.10 & 60.83 & 78.03 & 85.03 \\
        Formula         & 72.12 & 73.92 & 58.53 & 76.64 & 87.01 \\
        DP              & 47.71 & 60.98 & 47.00 & 76.39 & 80.24 \\
        \midrule
        \multicolumn{6}{c}{\textit{\ourmethodcot}} \\
        \midrule
        Upper Bound & \textcolor{gray}{90.17} & \textcolor{gray}{91.32} & \textcolor{gray}{66.82} & \textcolor{gray}{88.13} & \textcolor{gray}{98.47} \\
        DP\&Formula     & 80.92 & 82.94 & 54.38 & 80.49 & 93.23 \\
        Formula         & 76.70 & 79.17 & 53.92 & 78.41 & 90.76 \\
        DP              & 77.53 & 81.33 & 52.07 & 80.93 & 92.69 \\
        \midrule
        \multicolumn{6}{c}{\textit{Base Model}} \\
        \midrule
        DP & 46.84 & 54.81 & 18.81 & 55.62 & 69.76 \\
        \bottomrule
    \end{tabular}
    }
    \caption{Ablation study for the generation modes of \ourmethodda and \ourmethodcot (Qwen2.5-Coder-7B). ``Upper Bound'' indicates that at least one output is correct. }
    \label{tab:ablation_merge}
\end{table}

\paragraph{Effectiveness of generation modes}
To evaluate the effectiveness of the two generation modes (Formula and DP) in our framework, we conduct the ablation study for \ourmethodda\ and \ourmethodcot\ (Qwen2.5-Coder-7B) in Table \ref{tab:ablation_merge}. For both variants, excluding either generation mode generally leads to a performance decline. Specifically, for \ourmethodda, directly generating Formulas significantly outperforms directly generating answers, with improvements of 24.41\% and 11.53\% on the complex reasoning subsets of WTQ and HiTab, respectively. The Formula mode can concisely express the semantic intent of a question with a small number of tokens, whereas the DP mode struggles to directly generate answers without reasoning. In efficiency-critical scenarios, using only the Formula mode can be a viable solution. For \ourmethodcot, the CoT-style paradigm substantially improves performance, but the Formula mode no longer has a clear advantage over the DP mode due to possible execution errors and the limited Formula support of the \texttt{formulas} toolkit, which cannot fully match the capabilities of the Excel engine. Nevertheless, the two modes remain complementary, and their combination via mix self-consistency still yields better overall performance.

\begin{figure}[t]
    \centering
    \begin{subfigure}[b]{0.9\linewidth}
    \includegraphics[width=\linewidth]{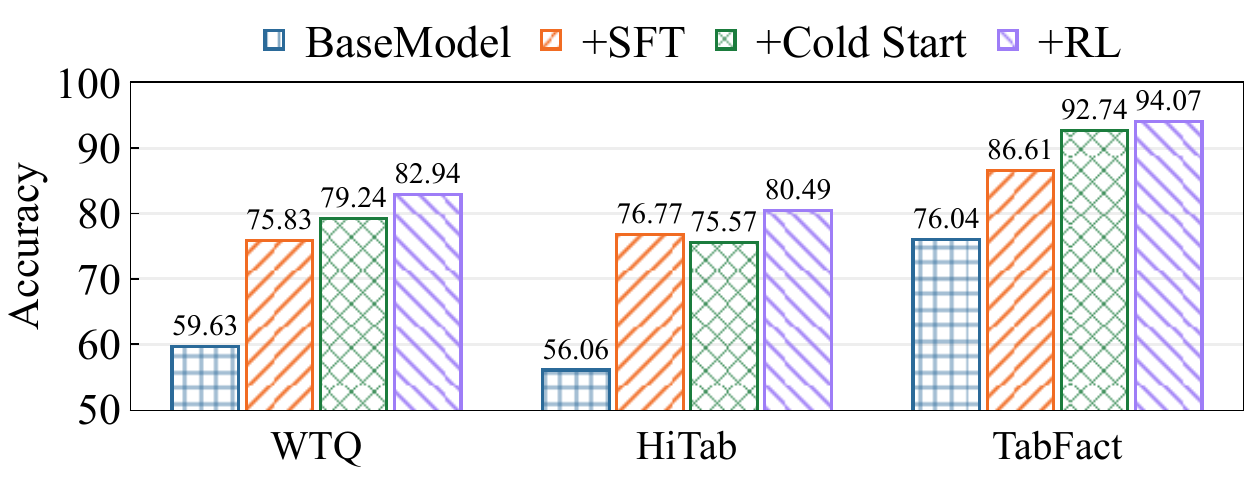}
    \caption{\ourmethodcot\ (Qwen2.5-Coder-7B)}
    \label{fig:merge_strategies_qwen2.5-7b}
    \end{subfigure}
    \begin{subfigure}[b]{0.9\linewidth}
    \includegraphics[width=\linewidth]{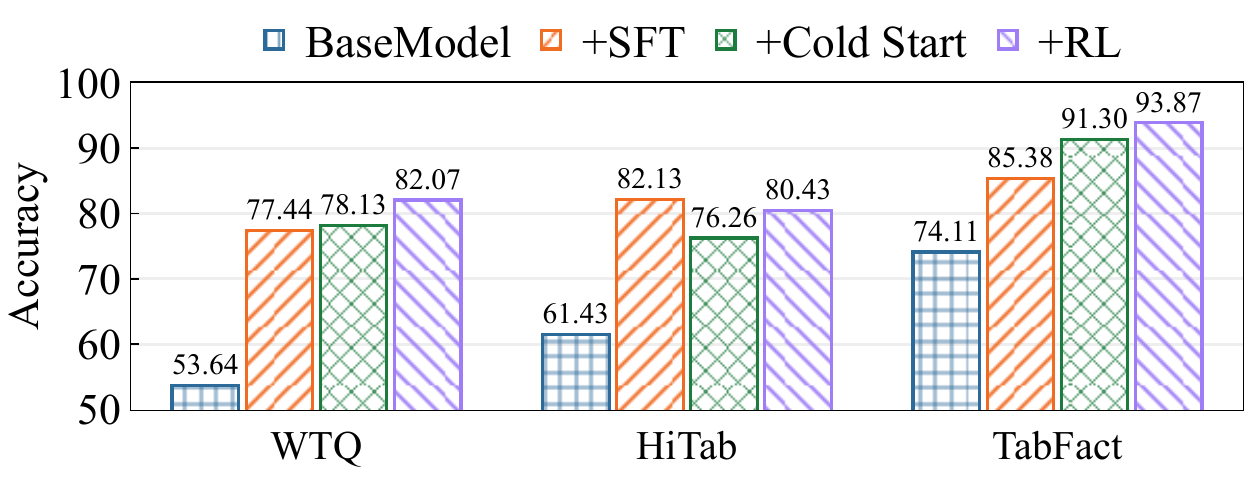}
    \caption{\ourmethodcot\ (Llama3.1-8B)}
    \label{fig:merge_strategies_llama3.1-70b}
    \end{subfigure}
    \caption{Ablation study for the training stages of \ourmethodcot (Qwen2.5-Coder-7B and Llama3.1-8B).}
    \label{fig:merge_strategies}
\end{figure}

\paragraph{Effectiveness of training stages}
To analyze the effectiveness of the training stages, we conduct the ablation study for \ourmethodcot\ (Qwen2.5-Coder-7B and Llama3.1-8B) in Figure \ref{fig:merge_strategies}. Across all models in this study, both DP and Formula modes are jointly used for generation, with the final answer also determined by mix self-consistency.
Before training, the base models show limited ability in TableQA. As the training progresses, the performance improves step by step. The SFT stage using a large-scale training dataset brings the most significant improvement, where both Formula generation and table understanding capabilities are greatly enhanced. During the Cold Start stage, we guide the model to perform reasoning before generating answers or Formulas. On WTQ and TabFact, this further unleashes the reasoning potential of LLMs, while on HiTab, performance drops slightly, possibly due to the limited size of cold-start data, which leads the model to overthink the simple questions in HiTab. 
The RL training stage enables further performance improvement by encouraging the model to explore and reinforcing successful reasoning through rewards.

\subsection{Additional Analysis}\label{sec:additional_analysis}

\begin{figure}[t]
    \centering
    \includegraphics[width=\linewidth]{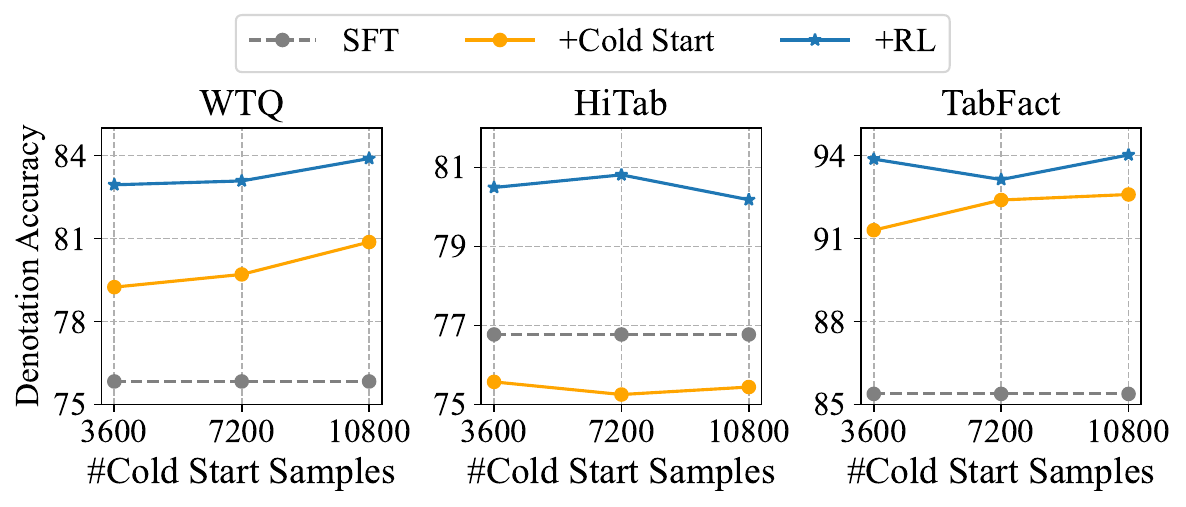}
    \caption{Impact of cold start samples for \ourmethodcot (Qwen2.5-Coder-7B).}
    \label{fig:cold_start_exp}
\end{figure}

\paragraph{Impact of Cold Start Samples}
In Figure \ref{fig:cold_start_exp}, we compare the impact of different amounts of cold start samples on \ourmethodcot\ (Qwen2.5-Coder-7B). In previous experiments, we used 3,600 samples from \ourdatasetcot as the cold start data before RL training. Here, we additionally annotate 7,200 more cold start samples using Claude-Sonnet-4, resulting in a total of 10,800 samples. We then conduct experiments under three different cold start data scales: 3,600, 7,200, and 10,800 samples, to examine whether more cold start data leads to improved performance.
On WTQ, both the Cold Start and corresponding RL performance improve as the number of cold start samples increases, possibly because the questions in WTQ rely more heavily on the reasoning process.
On HiTab and TabFact, \ourmethodcot does not benefit from additional CoT data. This may be because the data in these two datasets are relatively homogeneous and easy to handle by Formula or DP mode, making the performance differences after RL training more attributable to differences in training trajectories.

%% file: 6-Conclusion.tex
\section{Conclusion}
In this paper, we construct \ourdataset, a Formula-annotated dataset to enhance the Spreadsheet Formula generation capabilities of LLMs for TableQA. We propose \ourmethod, an answer-Formula joint reasoning framework that leverages the complementarity of Formula and Direct Prompt (DP) in solving lookup questions and complex reasoning questions, respectively. The extensive experimental results on the WTQ, HiTab, and TabFact datasets demonstrate that both variants of \ourmethod\ consistently outperform existing fine-tuning-based baselines. \ourmethodda achieves a favorable balance between reasoning efficiency and accuracy, while \ourmethodcot further achieves higher performance by explicit CoT reasoning. Additionally, empirical studies on the out-of-domain datasets like AIT-QA and FinQA demonstrate the generalizability of \ourmethod\ in new domains and the HybridQA task. These facts confirm the effectiveness and versatility of \ourmethod\ in enhancing the table reasoning capabilities of LLMs.

%% file: 8-Appendix.tex
\onecolumn

\section{A Details for converting SQL queries into Formulas}\label{sec:sql2formula}
We developed the Formula conversion program based on the SQL queries and templates annotated by Squall.  To handle a broader range of questions, Squall standardizes the WTQ tables and introduces two special columns: ``id'' and ``agg''. The ``id'' column indicates the row index, while the ``agg'' column indicates whether a row is a ``Total'' summary row. By applying special conversion rules to the ``id'' and ``agg'' columns, we can further simplify formula expressions. \textbf{For entities that can be looked up from the table, we directly use their coordinates (e.g. ``\texttt{A5}'') instead of using operations like ``\texttt{FILTER}'' to make the Formulas more concise.} For example, for the SQL ``SELECT c1 FROM w ORDER BY id ASC LIMIT 1'', the complex reasoning formula ``INDEX(A2:A13, MATCH(MIN(D2:D13), D2:D13, 0))'' can be simplified to the lookup formula ``=A1'' by interpreting the ``id'' column semantics.

\section{B Criteria for Dividing the Complex Subset}\label{sec:subset_criteria}
To demonstrate the ability of our method in performing complex reasoning, we divide the complex subsets from WTQ and HiTab for evaluation.
Since the WTQ dataset only contains annotations for answers, we used GPT-4o to 
divide the complex reasoning subset. We prompt the LLM to determine whether answering the questions requires arithmetic operations, date calculations, value aggregation, or other operations that may involve numbers. As a result, we divide the WTQ complex reasoning subset containing 2,421 samples.
Since HiTab annotates Formulas, we classify Formulas other than the assignment Formulas as the complex reasoning subset of HiTab, as these answers cannot be directly obtained from tables. The HiTab complex reasoning subset consists of 217 samples. 

\section{C Experimental Hyperparameters}\label{sec:hyperparameters}
We mainly apply \ourmethod to two widely used LLMs: Qwen2.5-Coder-7B \cite{hui2024qwen2}, and Llama3.1-8B \cite{dubey2024llama}. 
During supervised fine-tuning (SFT) for \ourmethodda, we adopt LoRA \cite{hu2022lora} with a cosine annealing scheduler, using an initial learning rate of $2 \times 10^{-4}$, batch size of 8, and a maximum sequence length of 16,384. Models are trained for 3 epochs on \ourdatasetda.
During cold-start SFT for \ourmethodcot, we switch to full-parameter fine-tuning, set the learning rate to $7 \times 10^{-6}$, and continue training 3 epochs using \ourdatasetcot based on \ourmethodda.
During RL training for \ourmethodcot, we employ the GRPO algorithm with a prompt length of 16,384, response length capped at 2,048, mini-batch size of 256, a rollout size of 10, and a learning rate of $1 \times 10^{-6}$. The KL divergence coefficient is set to 0 to enhance training stability. This stage uses the original WTQ, HiTab, and TabFact datasets over 3 training epochs.

\section{D The Prompts for LLMs}\label{sec:prompt}
In this section, we provide the prompts used to annotate Formulas for \ourdataset\ construction and the prompts to instruct LLMs to generate Formulas or direct answers. 

\subsection{D.1 Prompt for Annotating \ourdatasetda}\label{sec:prompt_annotate}
We utilize the prompt to annotate pure Formulas for the complex questions in WTQ and the randomly sampled questions in TabFact. For each example, we sample 10 responses to check whether any of the generated Formulas produce an execution result that matches the gold answer. If none match, we perform additional rounds of sampling, up to a maximum of three rounds in total. If a match is found, one of the matching responses is selected and added to \ourdatasetda.

\subsubsection{First Annotation Round} 

\begin{promptbox}
\textbf{\underline{System: }} \\
Given an Excel table, a question, and the answer to the question. Write an Excel formula (=???) to obtain the answer. You should only output the formula.\\
\textbf{\underline{User: }} \\
\textbf{[Table Title]}\{table\_title\}\\
\textbf{[Table]}\{table\_str\}\\
\textbf{[Question]} \{question\}\\
\textbf{[Gold Answer]} \{gold\_answer\}\\
\textbf{[The Formula You think to answer the question]} \\
\textbf{\underline{Assistant: }} \\
\# Generated Formulas
\end{promptbox}

\subsubsection{Subsequent Annotation Rounds}

\begin{promptbox}
\textbf{\underline{System: }} \\
Given an Excel table, a question, and the answer to the question. Write an Excel formula (=???) to obtain the answer. You should only output the formula.\\
\textbf{\underline{User: }} \\
\textbf{[Table Title]}\{table\_title\}\\
\textbf{[Table]}\{table\_str\}\\
\textbf{[Question]} \{question\}\\
\textbf{[Gold Answer]} \{gold\_answer\}\\
In the previous rounds, you attempted to generate the following formulas, but none of them is correct. Please try to generate the correct formula. You should only output the Formula. \\
\textbf{[The wrong formula you generate in previous rounds]}\\
1. Formula is: \{formula\} ; Execution Results is: \{result\} \\
2. Formula is: \{formula\} ; Execution Results is: \{result\} \\
...\\
\textbf{[The Formula You think to answer the question]}\\
\\
\textbf{\underline{Assistant: }} \\
\# Generated Formulas
\end{promptbox}

\subsection{D.2 Prompt for Annotating \ourdatasetcot}\label{sec:prompt_annotate_cot}
We utilize the prompt to annotate Formulas or Answers with CoT reasoning responses for randomly sampled questions in WTQ, TabFact, and HiTab. For each dataset and each generation mode, we annotate 1,800 CoT examples and verify their correctness using the gold answers. All the data collectively form \ourmethodcot, totaling 10,800 examples.

\subsubsection{Annotating Formulas with CoT}

\begin{promptbox}
\textbf{\underline{System: }} \\
You are an advanced Table-question solving assistant with access to the Excel Formula Engine Tool. Given a question and the corresponding table, your task is to generate a formula to solve this question. I'll execute the generated formula to obtain the final answer. 

\#\# Response Structure
Your response must follow this specific format:
1. \texttt{<think> ...think with less than 300 words ... </think>\textbackslash n\textbackslash n<formula>=...</formula>}
2. If there are multiple formulas, join them with "$|$" (e.g. "=MAX(A1:A10);=MIN(B2:B5)" ).\\
\textbf{\underline{User: }} \\
\textbf{[Table Title]}\{table\_title\}\\
\textbf{[Table]}\{table\_str\}\\
\textbf{[Question]} \{question\}\\
\textbf{[The Formula You think to answer the question]} \\
\textbf{\underline{Assistant: }} \\
\# \texttt{<think>...<think>\textbackslash n\textbackslash n <formula>=...</formula>}
\end{promptbox}

\subsubsection{Annotating Answers with CoT}

\begin{promptbox}
\textbf{\underline{System: }} \\
You are an advanced Table-question solving assistant. Given a question and the corresponding table, your task is to analyze the question and generate the final answer. 

\#\# Response Structure
Your response must follow this specific format:
1. \texttt{<think> ...think with less than 300 words ... </think>\\n\\n<answer>The most concise answer</answer>}
2. Ensure that the content inside \texttt{`<answer>...</answer>`} is as concise as possible (e.g. "21 years" should be "21"). Additionally, If there are multiple answers, join them with "$|$" (e.g. "Tom|Carl|Lisa").\\
\textbf{\underline{User: }} \\
\textbf{[Table Title]}\{table\_title\}\\
\textbf{[Table]}\{table\_str\}\\
\textbf{[Question]} \{question\}\\
\textbf{[The Formula You think to answer the question]} \\
\textbf{\underline{Assistant: }} \\
\# \texttt{<think>...<think>\textbackslash n\textbackslash n <answer>...</answer>}
\end{promptbox}

\subsection{D.3 Prompt for \ourmethodda}\label{sec:prompt_da}
We utilize the prompt to guide the LLMs to directly generate Formulas or Answers. We provide the example of tabular data to show the difference between generating Formulas and generating answers. 

\subsubsection{Generating Direct Formulas} \label{sec:prompt_formula}

\begin{promptbox}
\textbf{\underline{System: }} \\
You are an Excel Expert. Based on the Excel table, generate excel formula to answer the user question.\\
\textbf{\underline{User: }} \\
\textbf{[Table Title]} agri-food industry sub-groups for workers aged 15 years and over, 2011\\
\textbf{[Table]}\\
$|0|A|B|C|D|E|$\\
$|1|sub-groups of agri-food|eastern ontario|eastern ontario|northern ontario|northern ontario|$\\
$|2|sub-groups of agri-food|french workers|other workers|french workers|other workers|$\\
$|3|sub-groups of agri-food|percent|percent|percent|percent|$\\
$|4|input and service supply|2.9|2.1|2.9|1.3|$\\
$|5|food, beverage, and tobacco processing|9.7|6.0|3.0|3.3|$\\
$|6|food retail and wholesale|35.3|31.3|39.1|37.3|$\\
$|7|food service|52.1|60.6|55.0|58.1|$\\
\textbf{[Question]}\\
\{question\}\\
\textbf{[Formula]}\\
\\
\textbf{\underline{Assistant: }} \\
\# Ouptut Formula
\end{promptbox}

\subsubsection{Generating Direct Answers} \label{sec:prompt_answer}

\begin{promptbox}
\textbf{\underline{System: }} \\
This is table question answering task, based on the given table, answer the given question. Only output the answer.\\
\textbf{\underline{User: }} \\
\textbf{[Table Title]} agri-food industry sub-groups for workers aged 15 years and over, 2011\\
\textbf{[Table]}\\
$|sub-groups of agri-food|eastern ontario|eastern ontario|northern ontario|northern ontario|$\\
$|sub-groups of agri-food|french workers|other workers|french workers|other workers|$\\
$|sub-groups of agri-food|percent|percent|percent|percent|$\\
$|input and service supply|2.9|2.1|2.9|1.3|$\\
$|food, beverage, and tobacco processing|9.7|6.0|3.0|3.3|$\\
$|food retail and wholesale|35.3|31.3|39.1|37.3|$\\
$|food service|52.1|60.6|55.0|58.1|$\\
\textbf{[Question]}\\
\{question\}\\
\textbf{[Answer]}\\
\\
\textbf{\underline{Assistant: }} \\
\# Ouptut Direct Answer
\end{promptbox}

\subsection{D.4 Prompt for \ourmethodcot}\label{sec:prompt_cot}
We utilize the prompt to guide the LLMs to generate Formulas or Answers output. We provide the example of tabular data to show the difference between generating Formulas and generating answers. 

\subsubsection{Generating Formulas with Cot Thinking} \label{sec:prompt_formula_cot}

\begin{promptbox}
\textbf{\underline{System: }} \\
You are an advanced Table-question solving assistant with access to the Excel Formula Engine Tool. Given a question and the corresponding table, your task is to generate a formula to solve this question. I'll execute the generated formula to obtain the final answer. 

\#\# Response Structure
Your response must follow this specific format:
1. \texttt{<think> ...think with less than 300 words ... </think>\textbackslash n\textbackslash n<formula>=...</formula>}
2. If there are multiple formulas, join them with "$|$" (e.g. "=MAX(A1:A10);=MIN(B2:B5)" ).\\
\textbf{\underline{User: }} \\
\textbf{[Table Title]} agri-food industry sub-groups for workers aged 15 years and over, 2011\\
\textbf{[Table]}\\
$|0|A|B|C|D|E|$\\
$|1|sub-groups of agri-food|eastern ontario|eastern ontario|northern ontario|northern ontario|$\\
$|2|sub-groups of agri-food|french workers|other workers|french workers|other workers|$\\
$|3|sub-groups of agri-food|percent|percent|percent|percent|$\\
$|4|input and service supply|2.9|2.1|2.9|1.3|$\\
$|5|food, beverage, and tobacco processing|9.7|6.0|3.0|3.3|$\\
$|6|food retail and wholesale|35.3|31.3|39.1|37.3|$\\
$|7|food service|52.1|60.6|55.0|58.1|$\\
\textbf{[Question]}\\
\{question\}\\
\textbf{[Formula]}\\
\\
\textbf{\underline{Assistant: }} \\
\# \texttt{<think>...<think>\textbackslash n\textbackslash n <formula>=...</formula>}
\end{promptbox}

\subsubsection{Generating Answers with Cot Thinking} \label{sec:prompt_answer_cot}

\begin{promptbox}
\textbf{\underline{System: }} \\
You are an advanced Table-question solving assistant. Given a question and the corresponding table, your task is to analyze the question and generate the final answer. 

\#\# Response Structure
Your response must follow this specific format:
1. \texttt{<think> ...think with less than 300 words ... </think>\\n\\n<answer>The most concise answer</answer>}
2. Ensure that the content inside \texttt{`<answer>...</answer>`} is as concise as possible (e.g. "21 years" should be "21"). Additionally, If there are multiple answers, join them with "$|$" (e.g. "Tom|Carl|Lisa").\\
\textbf{\underline{User: }} \\
\textbf{[Table Title]} agri-food industry sub-groups for workers aged 15 years and over, 2011\\
\textbf{[Table]}\\
$|sub-groups of agri-food|eastern ontario|eastern ontario|northern ontario|northern ontario|$\\
$|sub-groups of agri-food|french workers|other workers|french workers|other workers|$\\
$|sub-groups of agri-food|percent|percent|percent|percent|$\\
$|input and service supply|2.9|2.1|2.9|1.3|$\\
$|food, beverage, and tobacco processing|9.7|6.0|3.0|3.3|$\\
$|food retail and wholesale|35.3|31.3|39.1|37.3|$\\
$|food service|52.1|60.6|55.0|58.1|$\\
\textbf{[Question]}\\
\{question\}\\
\textbf{[Answer]}\\
\\
\textbf{\underline{Assistant: }} \\
\# \texttt{<think>...<think>\textbackslash n\textbackslash n <answer>...</answer>}
\end{promptbox}